# Medium Access using Distributed Reinforcement Learning for IoTs with Low-Complexity Wireless Transceivers

Hrishikesh Dutta and Subir Biswas

*Electrical and Computer Engineering, Michigan State University, East Lansing, MI, USA*
*duttahr1@msu.edu, sbiswas@msu.edu*

***Abstract* - This paper proposes a distributed Reinforcement Learning based protocol synthesis framework that can be used for synthesizing MAC layer protocols in wireless networks. The proposed framework does not rely on complex hardware capabilities such as carrier sensing and its associated algorithmic complexities that are often not supported in wireless transceivers of low-cost and low-energy IoT devices. Network protocols are first formulated as Markov Decision Processes (MDP) and then solved using reinforcement learning. A distributed and multi-Agent RL framework is used as the basis for protocol synthesis. Distributed behavior makes the nodes independently learn optimal transmission strategies without having to rely on full network level information and direct knowledge of the other nodes' behavior. The nodes learn to minimize packet collisions such that optimal throughput can be attained and maintained for loading conditions that are higher than what the known benchmark protocols such as ALOHA can support. In addition, the nodes are observed to be able to learn to act optimally in the presence of heterogeneous loading and network topological conditions. Finally, the proposed learning approach allows the wireless bandwidth to be fairly distributed among network nodes in a way that is not dependent on those heterogeneities. Via extensive simulation experiments, the paper demonstrates the performance of the learning paradigm and its abilities to make nodes adapt their optimal transmission strategies on the fly in response to various network dynamics.**

## I. INTRODUCTION

Protocol design in wireless networks is mostly driven by heuristics and past experience of network designers. The best practice for programming wireless MAC logic is to use known standardized protocols, including, ALOHA, CSMA, CSMA-CA, and their higher derivatives, such as WiFi, Bluetooth, and Zigbee. The specific family of protocol is chosen for an application based on the cost budget and lower layer hardware capabilities for carrier sensing, collision detection, and also depending on the network properties such as round-trip propagation delay. A notable drawback of this design approach based on available standardized protocols is that the resulting logic often underperforms in application-specific non-standard scenarios caused by topology- and load-heterogeneity, and other factors. For example, with baseline ALOHA MAC, the throughput of a network starts falling at higher loads due to collisions. The situation is compounded in the presence of heterogeneity. In a network with arbitrary mesh topology, some nodes may be in more disadvantageous position as compared to others in terms of the collisions they experience. As a result, those nodes receive less share of the available wireless bandwidth. The root cause of these limitations is the fact that the nodes in the network are statically programmed with standardized protocol logic and they lack the flexibility that can result in abilities to learn optimal behavior in specific scenarios.

The goal of this work is to develop a MAC layer protocol based on reinforcement learning that can address these challenges. With the long-term objective of making the learning approach generalized, we start with an initial scenario of a network with its nodes running the simplest MAC logic without relying on complex and energy-expensive lower-level capabilities such as carrier sensing. The developed framework would be useful for simple transceivers used in low-cost IoT devices and wireless sensor nodes.

The key idea behind distributed Reinforcement Learning (RL) for protocol design [1] is to formulate the protocol layer logic in each network node as a Markov Decision Process (MDP) and use RL as a temporal difference method to solve the MDP. The solution of this MDP is a set of transmission actions taken by individual nodes, thus resulting a network protocol. Since the RL mechanisms learn online, the need for prior data collection and training is avoided in this approach. However, in distributed RL, a key requirement is that an agent (i.e., a network node) needs to have as much network-level information as possible in order to avoid collisions and share wireless bandwidth in a fair manner. This becomes a problem in a partially connected network in which the information availability to a node is usually limited only to its immediate neighborhood.

An obvious solution to this problem is to employ centralized learning in which a centralized agent, with access to complete network level information, can learn optimal node behavior (i.e., a protocol) and downloads it to the individual nodes. However, this approach may not be feasible in decentralized practical scenarios. In this paper, we develop a distributed RL-based protocol synthesis approach that can work with partial/local network information availability in partially connected networks. The approach is built on our prior work [1] which assumed full network-level information availability for fully connected networks. In this work, the distributed learning mechanism is developed such that the RL agents (i.e., network nodes) rely only on the localized network information available in their immediate neighborhoods. It is shown the proposed mechanism allows nodes to learn to reduce collisions so as to mimic the baseline ALOHA behavior for low network traffic, reach the maximum ALOHA throughput, and unlike baseline ALOHA, maintain that maximum value for higher traffic loads. An important feature of the developed framework is that the wireless bandwidth is distributed in a fair manner irrespective of the underlying network load topology, even when they are heterogenous.

This work has the following scope and contributions. First, a distributed learning framework for synthesizing MAC layer protocol with localized network information visibility is designed. Second, it is shown that the synthesized protocol can obtain and maintain known benchmark throughputs of standard protocols (ALOHA in this case). Third, the proposed learning is shown to work for arbitrary mesh network topologies. Finally, it is shown that distributed learning can handle heterogeneity both in terms of load and network topology.

The rest of the paper is organized as follows. In section II, the network models used in this work is presented. In section III, the MAC logic synthesis is posed as an MDP problem, and RL is introduced as a viable solution approach. In Section IV, the proposed RL framework and all its components are discussed in detail. Section V presents all experimental results and their interpretations. In Section VI, the existing related works are reviewed. Finally, the work is summarized, and conclusions are drawn in section VI.

## II. Network and Traffic Model

Arbitrary mesh topologies of low-cost sensor nodes without carrier-sensing abilities are considered. Fig. 1(a) depicts the generalized topology representation in which a solid line represents the physical node connectivity, and the dashed lines indicate the data paths. For example, nodes $i$ and $k$ transmits packets to node $j$, and node $j$ transmits packets to nodes $i$ and $k$. The MAC layer traffic load model is created such that all packets generated in a node is sent to its 1-hop neighbors following a uniform random distribution. In other words, if a node $i$ has $K$ one-hop neighbors and its MAC layer load is $g_i$ Erlang following Poisson distribution, node $i$ sends $\frac{g_i}{K}$ Erlangs amount of traffic to each of its one-hop neighbors. The objective of the proposed learning mechanism is to make the nodes learn transmission strategies that can obtain and maintain optimal (and fair) distribution of wireless bandwidth by reducing packet collisions across heterogenous topologies and loading conditions.

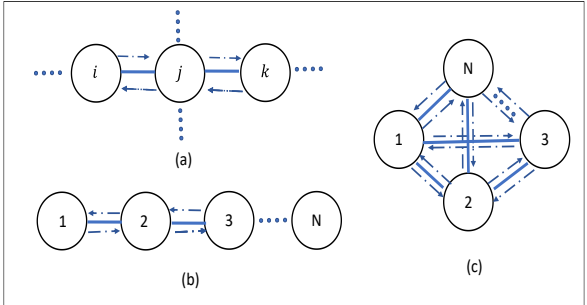

Fig. 1. Network and data sharing model for (a) generalized, (b) partially connected and (c) fully connected topology.

Fig. 1(b) and 1(c) are the minimally connected and fully connected examples of the generalized topology depicted in Fig. 1(a). In 1(b) nodes send data only two of its 1-hop neighbors, whereas in 1(c) a node sends data to all other nodes in the network. As described in Section IV, the network model includes the availability of piggybacking for sending very low data-rate control information using the data packets themselves. Such control information is used for local information sharing needed by the Reinforcement Learning mechanism.

## III. Protocol Synthesis as a Markov Decision Process

A Markov Decision Process (MDP) is used for modelling discrete stage sequential decision making in a stochastic environment [2]. In an MDP, the system transition takes place stochastically among a set of states $S = \{S_1, S_2, S_3, \dots, S_N\}$. State transitions take place as a result of an agent's actions defined in the action space $A = \{A_1, A_2, \dots, A_M\}$. The probability of transition from one state to another because of a specific action is defined by a set of transition probabilities $T$. For each action taken in a particular state, there is a numerical reward associated, which defines the physical benefit of taking the action in that state. An important property of MDP is that the probability of transition from the current state to the next state is dependent only on the current state and action. Hence the attribution Markov. Formally, an MDP can be defined by the tuple $(S, A, T, R)$, where S is the set of all possible states, A is the set of actions, T is the set of transition probabilities and R is the reward function. Thus, for any dynamic system, whose behavior can be modelled by an MDP, there exists a set of optimal actions for each state that would yield the maximum expected long-term reward [3].

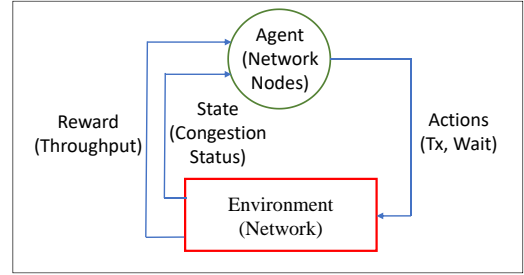

Fig. 2. Reinforcement Learning for Network Protocol MDP

Reinforcement Learning (RL) [4] is a class of algorithms for solving an MDP. The advantage of reinforcement learning over other MDP solving techniques (e.g., dynamic programming, etc.) is that RL does not always require an exact mathematical model of the system. One such model-free RL approach is to deploy Q-learning [5], in which the agent learns to take an optimal set of actions by repeatedly taking all the possible action in each state and evaluating their rewards. The best set of actions will be the ones that will maximize the expected long-term reward. During the learning process, the agents maintain a Q-table, whose entries are Q-values for each state-action pair. For a specific state, the action with the highest Q-value from the table is preferred by the agent. At the end of each decision epoch, the Q values for state-action pair $(s, a)$ are updated as shown in Eq. (1), where $r, \alpha, \gamma$ and $s'$ indicate the reward received, learning rate, discount factor, and the next state caused by action $a$, respectively.

$$Q(s,a) \leftarrow Q(s,a) + \alpha[r(s,a) + \gamma \times \max_{\forall a' \in A} Q(s',a') - Q(s,a)] \tag{1}$$

The dynamic behavior of a wireless network in different operating conditions can be represented as a Markov Decision Process. An annotation of different components of an MDP, representing a wireless network protocol, is shown in Fig. 2. In this formulation, the nodes of the network represent distributed agents that individually take actions to transmit with various probabilities or to wait. The environment with which an agent/node interact is the wireless network itself. The congestion levels in the network represent individual agent perceivable state space of the environment, which changes depending on the independent agents' actions. The reward can be a function of different performance metrics (such as, throughput, energy efficiency etc.) that need to be optimized.

## IV. DISTRIBUTED RL WITH LOCALIZED INFORMATION

The developed learning framework is termed as Distributed RL with Localized Information (DRLI) for synthesizing wireless MAC behavior. With DRLI, each wireless node acts as an independent learning agent and implements a specific flavor of learning, namely, Hysteretic Q-Learning [6]. A specific feature of DRLI is that unlike many prior work [1, 7], the learning agents/nodes do not rely on global network information. Instead, they use localized network information from the immediate neighborhood for learning effective MAC layer behavior in terms of transmission policies.

### A. Hysteretic Q-Learning

It is a distributed Q-learning algorithm used in a multi-agent co-operative setting to achieve a specific goal as a team. In Hysteretic Q-Learning, the agents do not participate in explicit communication and are unaware of each other's actions during the learning process. The primary difference of this approach as compared with traditional Q-learning is in its use of two learning rates $\alpha$ and $\beta$. The Q-table update rule are:

$$\delta = r + (\gamma) \times \max_{\forall a' \in A} Q(s', a') - Q(s, a)$$

$$Q(s, a) \leftarrow \begin{cases} Q(s, a) + \alpha \times \delta, & if\ \delta \geq 0 \\ Q(s, a) + \beta \times \delta, else \end{cases} \quad (2)$$

The parameter $\delta$ in Eq (2) controls which learning rate is to be used in the Q-table update equation in a particular epoch. In a multi-agent setting, an agent may be penalized even for an optimal action, if the actions taken by the other agents in the team were bad. Thus, the rewards and penalties received are dependent not only on their own actions, but also on other agents' actions. That is the reason behind the usage of two learning rates in Hysteretic Q-Learning. As can be seen from the equation, a positive value of the parameter $\delta$ leads to the use of $\alpha$ as the learning rate and a negative value of $\delta$ makes the algorithm use $\beta$ as the learning rate, where $\alpha$ and $\beta$ are the increase and decrease rates of the Q values respectively. The parameter $\delta$ is positive if the actions taken were beneficial for attaining the desired optimum of the system and vice-versa. Hence, $\beta$ is chosen such that it is always less than $\alpha$ in order to assign less importance to the penalties.

In DRLI learning framework, nodes act as hysteretic learning agents, and iteratively learn their individual optimal transmission strategies in a distributive manner. The objective is to maximize the network performance and fairness.

### B. Learning with Localized Network Information

In a cooperative setting, where a team of learning agents share a common objective, solving an MDP in a distributed fashion is easier with global information availability. However, in the context of a wireless network, it is not always feasible for an RL agent (i.e., a node) to have global information such as network wide congestion and throughput. The problem is particularly compounded for partially connected mesh topologies.

In the proposed learning framework, a reward function for an agent/node is designed such that information only from its one-hop neighbor nodes is used. Such information is gathered using an in-band mechanism [8] using which the relevant control information is piggybacked over regular MAC layer PDUs. Consider a scenario in which $s_j$ is the current throughput of node-$j$, and $s_{j \to i}$ is the part of $s_j$ for which node-$i$ (i.e., a 1-hop neighbor of node-$j$) is the intended receiver. Node-$i$ computes the quantity $s_{j \to i}$ by keeping track of the non-collided transmissions from node $j$ that are intended for node $i$. Node-$i$ periodically sends $s_{j \to i}$ information to node-$j$ by piggybacking it within its wireless MAC PDUs. Node $j$ periodically computes its own overall throughput $s_j = \sum_{\forall i} s_{j \to i}$, where $i$ represents all its 1-hop neighbors. Node $j$ also periodically sends its own recently computed throughput $s_j$ to all its 1-hop neighbors using piggybacking. A typical MAC layer PDU from node-$i$ is shown in Fig. 3 below.

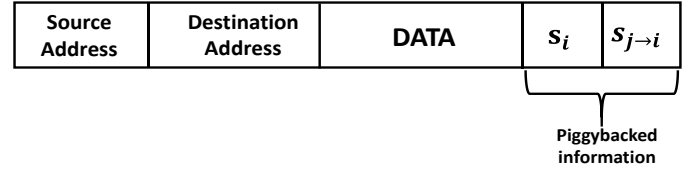

Fig. 3: MAC layer PDU from node $i$

The above process is periodically executed across all 1-hop neighbors. The process creates an information distribution model such that at any given point in time, each network node-$j$ possesses the throughput information of all its one-hop neighbors. As explained in the following section, the RL reward function is developed using this 1-hop neighborhood throughput information.

It is important to note that the lack of global network information visibility is different from the concept of Partially Observable MDP (POMDP [9, 10]). In POMDP, the agents lack visibility to their current states, and they infer the states based on a set of observations after the actions [11]. In this work, on the other hand, an agent/node has visibility about its own state, which is defined by the 1-hop congestion status. It is just that the lack of global information visibility forces the state to be defined based on local information, which can have performance implications for the MDP solutions.

### C. Distributed Reinforcement Learning Components

Each node runs an RL agent, which is used for solving its own MDP based on local information model as described in Section IV B. Following are the primary components of the proposed distributed RL based learning framework.

Action space: The actions are defined as probabilities of packet transmissions. In order to keep the action space discrete, the transmission probability is divided into 20 equal steps in the range $[0, 1]$. Formally stated, an action $a \in \{a_1, a_2, \dots \dots, a_{20}\}$ represents the probability with which a packet is transmitted following the Hysteretic RL logic as explained in Section IV A.

State space: The state space is defined as the network congestion level estimated from packet collision status. The state perceived by an RL agent running in a node at an RL epoch is defined as the probability of packet collisions experienced by the node during that epoch. As done for the action space, the states are also discretized in 24 equal steps in the range [0, 1]. Hence, the state of a node $i$ at an epoch $t$ is denoted as $\hat{s}_i(t) \in \hat{S}_i = \{\hat{s}_{i,1}, \hat{s}_{i,2}, \dots \dots, \hat{s}_{i,24}\}$, where $\hat{S}_i$ is the state space for node $i$. Each state $s_{i,k}$ represents a level of collision observed by node-$i$. The collision level is quantified as:

$$P_c = \frac{Number\ of\ packet\ collisions}{Number\ of\ transmitted\ packets} \quad (3)$$

<u>Rewards</u>: The observables of a node $i$ is defined as the information about its own throughput $s_i$ and its one-hop neighbors' throughput $s_j$ (for all neighbors $j$), where $s_i$ and $s_j$ are expressed in packets/packet duration, or Erlang. This throughput information is exchanged using piggybacking as explained in *Subsection A*.

Table I: Reward structure formulation

| # | $\Delta s_i - \epsilon$ | $\Delta f_i$ | $R_i$ |
|---|---|---|---|
| 1 | + | + | +50 |
| 2 | + | - | -30 |
| 3 | - | + | +10 |
| 4 | - | - | -50 |

We define a temporally sensitive reward structure as shown in Table I. Here $\Delta s_i = s_i(t) - s_i(t-1)$, which is the discrete time derivative of throughput, and $\Delta f_i = f_i(t) - f_i(t-1)$ is the time derivative of throughput fairness. Fairness is defined as $f_i(t) = -\sum_{\forall j \neq i} |s_i(t) - s_j(t)|$ such that a larger $f_i$ indicates a fairer system. If the gradients of throughput and fairness over time are positive, which are desirable trends, a high positive reward is awarded to the RL agent. The reverse is done when the gradients are negative. Instead of using throughput gradient directly, we use the gradient of $\Delta s_i - \epsilon$, where a small positive quantity (i.e., $\epsilon = 0.005$) is used to reduce oscillations post convergence. The reward assignment is designed in such a way that the nodes learn to maximize their individual throughput while preserving fairness among their directly connected neighbors. In addition to the baseline rewards from Table I, a learning recovery protection is provided by assigning a penalty (i.e., 0.8) to an agent if its individual throughput $s$ ever goes to zero. More about the design hyper-parameters are presented in Section VI.

## V. Experiments and Results

The performance of DRLI-enabled MAC (DRLI-MAC) is evaluated in networks with arbitrary mesh topology. Its performance is then compared with a benchmark protocol with known-best performance in the absence of carrier sensing and network time-synchronization (i.e., slotting) capabilities. The learning hyperparameters for DRLI-MAC are shown in Table II. The experiments are conducted using a *C*-language based simulator.

Table II: Baseline Experimental Parameters

| No. of packets per epoch | 1000 |
|---|---|
| $\alpha$ | 0.9 |
| $\beta$ | 0.1 |
| $\gamma$ | 0.95 |
| $\epsilon$ | $1.0 \times e^{-epoch\ ID/1000}$ |

### A. Performance in a 3-Node Network

To understand the operations of the learning paradigm in a simplified network, we first simulate the DRLI-MAC logic in a 3-node partially connected topology. As shown in Fig 4(a), throughput for DRLI-MAC under homogeneous load closely matches with that of the protocol with known best performance, which is unslotted ALOHA [12]. ALOHA throughput attains the maximum (i.e., $s_1 = s_2 = s_3 \approx 0.08$), when the optimum loads $\widehat{g_1} = \widehat{g_2} = \widehat{g_3} \approx 0.2$, and then decreases asymptotically with increasing load. The reduction is due to increased packet collisions. With DRLI-MAC, the RL agents in nodes learn to adjust the transmit probabilities such that their behaviors mimic that of ALOHA at lower loads (i.e., $g < \hat{g}$), and outperforms at higher loads by limiting collisions and maintaining the maximum throughput. Figs. 4(b) and (c) depict the convergence behavior of node-2 over different learning epochs. The behavior is shown for a specific load distribution $g_1 = g_2 = g_3 = 0.75$. It is observed that node-2 learns to take the optimum action with action ID=5 so that its effective node-specific load exerted to the network (i.e., $g_2^*$) converges to the optimum load $\widehat{g_2} \approx 0.2$. This makes sure that the collisions in the network are reduced and the optimal throughput is maintained even at higher network traffic. Similar convergence behaviors to attain optimal actions and effective loads were also observed for nodes 1 and 3.

An important observation from Fig 4(a) is that the per-node throughput is fairly distributed among all network nodes. This is more evident in heterogeneous traffic conditions as shown in Fig. 5. With ALOHA access strategy, there is a high variation of throughputs among the three nodes for heterogeneous load distribution. In contrast, with DRLI-MAC, the differences in throughputs of individual nodes are significantly smaller. In each of the three plots in Fig. 5, the loads from node-1 ($g_1$) and node-2 ($g_2$) are kept fixed at different values, and the node-level throughput variations are observed for varying load from node-3 ($g_3$). These represent the scenarios: $g_1 \le \hat{g}, g_2 \le \hat{g}, g_1 \le \hat{g}, g_2 > \hat{g}$ or $g_1 > \hat{g}, g_2 \le \hat{g}$, and $g_1 > \hat{g}, g_2 > \hat{g}$. It can be observed that with DRLI-MAC, the RL agents in nodes learn to adjust the transmit probability such that the available wireless bandwidth is fairly distributed. Also notable is the fact that the DRLI-MAC logic can hold the optimal throughput for higher network loads, even under heterogeneous loading conditions.

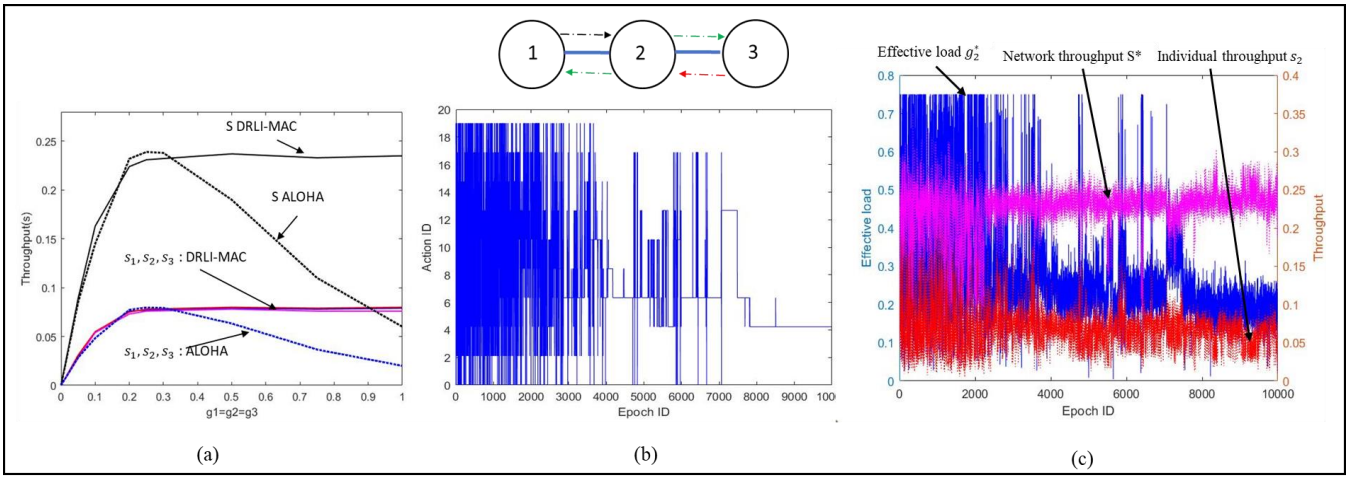

Fig. 4. (a) Load vs Throughput for 3-node linear network; $s_i$, $g_i$ are the individual throughputs and loads respectively for node $i$, $S$ is the network throughput, (b) Convergence plots for actions of node 2 and (c) Convergence plots for effective load and throughput of node 2 and network throughput

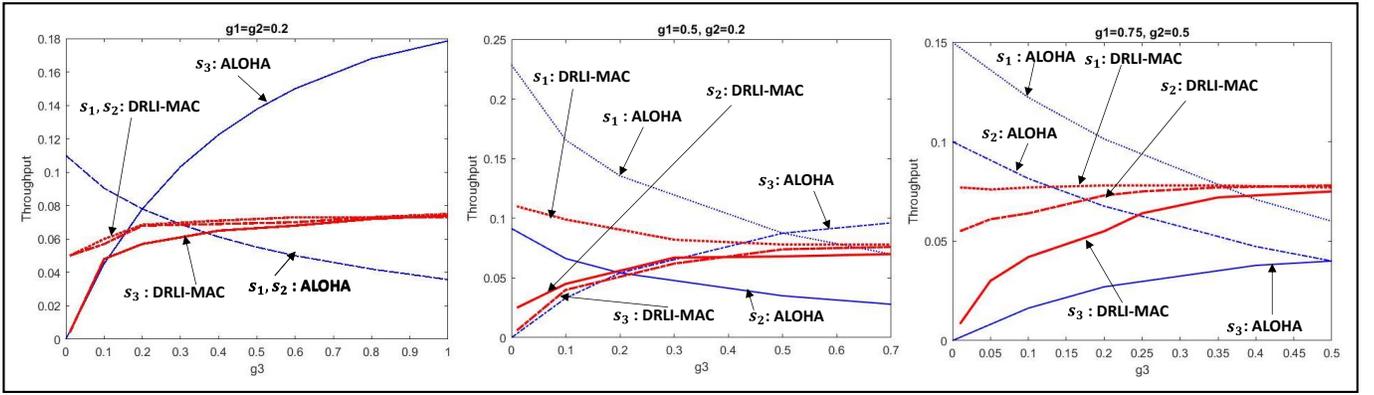

Fig. 5: Performance in a 3-node network with heterogeneous load

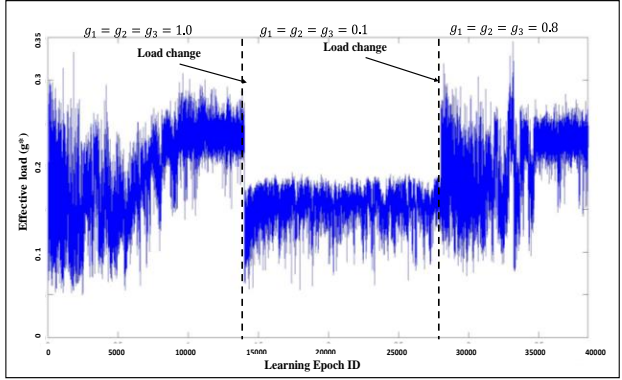

Fig. 6. Performance of DRLI-MAC in a dynamic environment

Fig. 6 demonstrates the online learning abilities of the underlying RL framework in the presence of dynamic loading conditions. In the figure, it is shown how the nodes learn to adjust their transmit probabilities with changes in the network load. Initially, the application layer load in the network is $g_1 = g_2 = g_3 = 1.0$, and the network throughput (S) converges to the optimal value of 0.24. There are changes in the incoming application layer load indicated by the dotted vertical lines, where the load changes to $g_1 = g_2 = g_3 = 0.1$ and $g_1 = g_2 = g_3 = 0.8$ respectively. It is observed that the RL framework allows the nodes to maintain the optimal throughputs even in these dynamic loading situations. This is achieved by the nodes learning to transmit packets with an optimal transmit probability according to the incoming application layer traffic.

These results indicate how the proposed learning framework is able to learn an effective wireless medium access logic in a distributed manner. Also, to be noted that the learning relies only on localized network information, thus rendering the technique highly scalable.

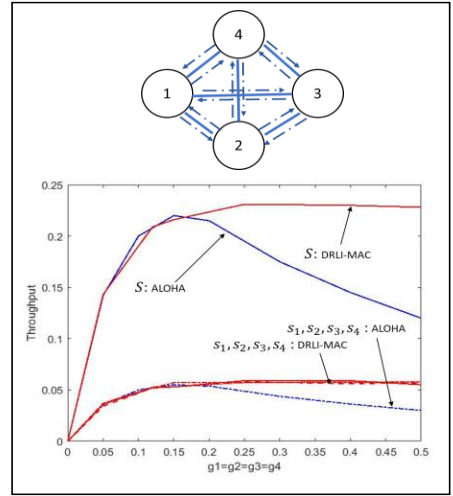

Fig. 7. Load-Throughput for 4-node fully connected network; $s_i, g_i$ are the throughputs and loads respectively for node $i$, $S$ is the network throughput.

### B. Performance in a 4-Node Network

**Fully connected topology**: In a fully connected network, the learning agents in each node has full access to the entire network level information. As shown in Fig. 7, such full information access allows the DRLI-MAC logic to attain the maximum possible throughput at the optimal loading condition. Also, unlike ALOHA, at higher loads, it is able to maintain that maximum throughput by learning to adjust the transmission probabilities in order to keep packet collisions in check. Moreover, the network throughput is fairly distributed among all four nodes. Learning and convergence behavior for this network was found to be following the same patterns for the 3-node network as observed in Fig. 4.

**Partially connected topology**: MAC learning logic was also applied in a partially connected 4-node topology as shown in Fig 8. Unlike in the topology in Fig. 7, information availability for a node in this case is restricted only within 1-hop neighborhood of the node.

Fig. 8(a) indicates the networkwide throughput variation for ALOHA with different combinations of $g_1(= g_4)$ and $g_2(= g_3)$. It can be seen that the network throughput is maximized when $g_2 = g_3 \to 0$. This is because, in this scenario, collisions experienced by the packets from the edge nodes 1 and 4 are negligible. However, this is not a desirable behavior, since in addition to maximizing network throughput, the objective is to distribute the throughput fairly. The dashed line in this 3-D plot indicates the situation in which the throughput is fairly distributed among the nodes. A projection of this line on the $g_1(=g_4)$ plane is shown in Fig. 8(b). From these plots, it can be observed that the desired behavior (maximum throughput with fair bandwidth distribution) is achieved when $g_1 = g_4 \approx 0.2$ and $g_2 = g_3 \approx 0.3$. The optimum individual node throughput in this situation is obtained as $\hat{s}_i = 0.058$.

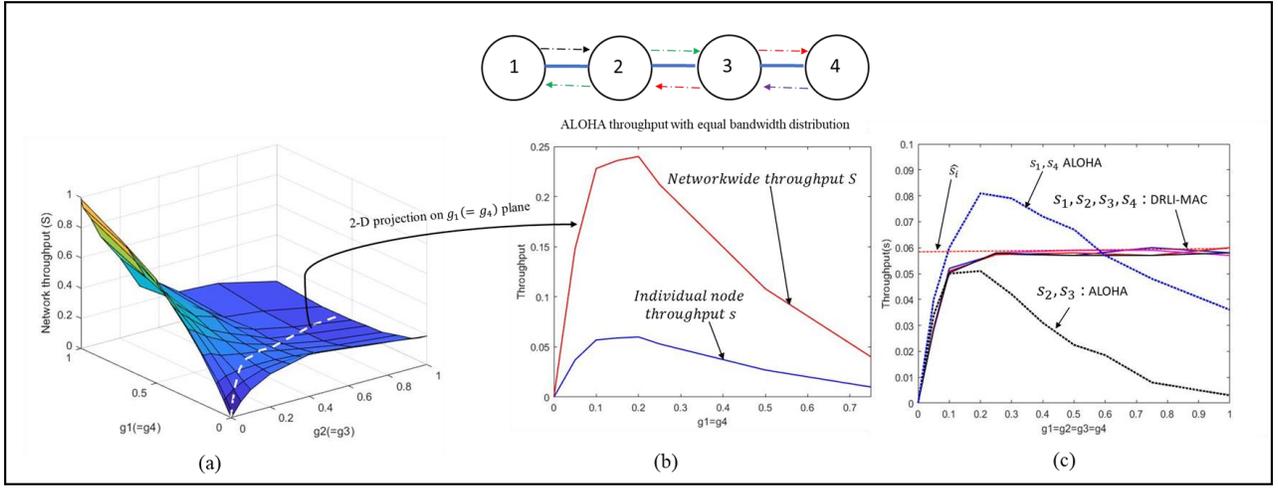

Fig. 8. (a) Surface plot indicating the network throughput variation of ALOHA with different $g_1(=g_4)$ and $g_2(=g_3)$ combinations, (b) ALOHA throughput with fair bandwidth distribution, (c) Load vs Throughput of DRLI-MAC for 4-node partially connected network; $s_i$, $g_i$ are the throughputs and loads respectively for node $i$, $S$ is the network throughput, $\hat{s}_i$ is the maximum ALOHA individual throughput with fair bandwidth distribution.

It can be observed from Fig. 8(c) that for regular ALOHA, the node throughputs not only reduce at higher loads, but they are also unfairly distributed. Because of higher exposure to collisions (i.e., due to higher topology degrees), nodes 2 and 3 have lower throughputs compared to nodes 1 and 3. In addition to addressing this fairness issue, DRLI-MAC is able to achieve the optimal throughput ($\hat{s}_i$) obtained by the known benchmark protocol ALOHA, as shown in Fig. 8(b). The RL framework also allows the nodes to hold the optimal throughput for higher traffic in the network.

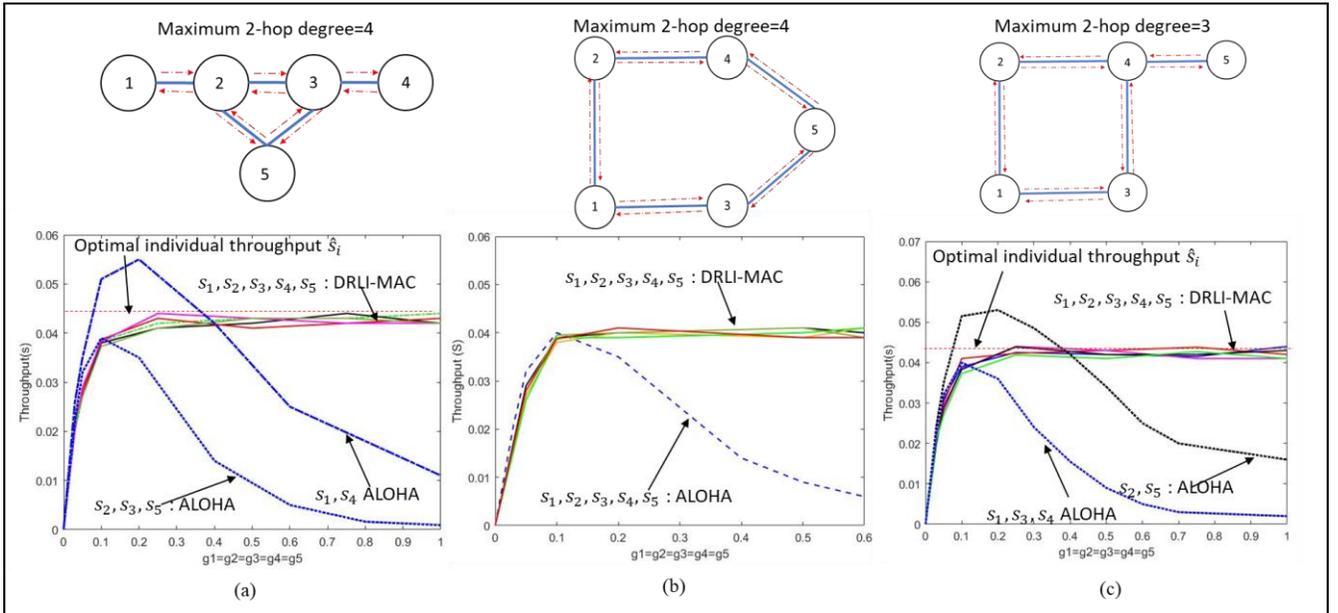

Fig. 10: Performance of DRLI-MAC in a five-node network of arbitrary topology

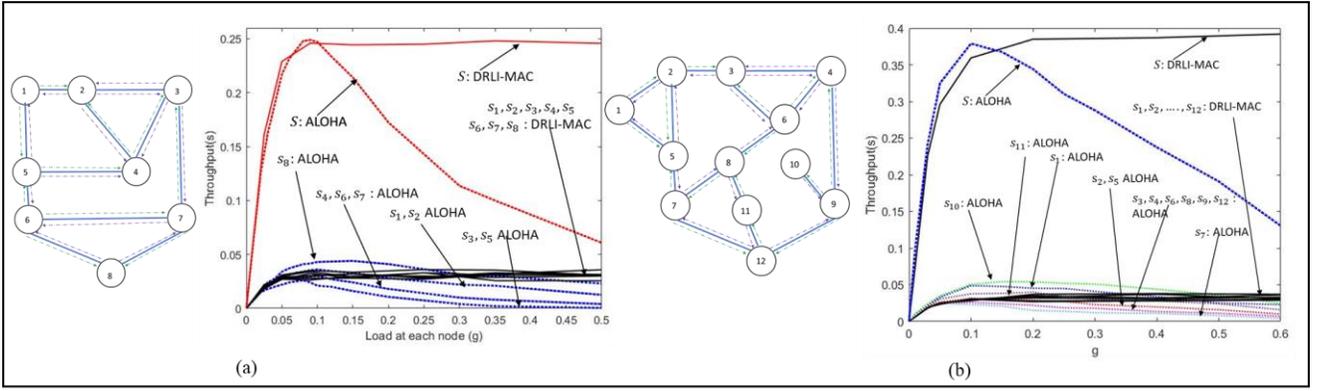

Fig. 11: Performance of DRLI-MAC in large networks with (a) 8 nodes and (b) 12 nodes of arbitrary topology

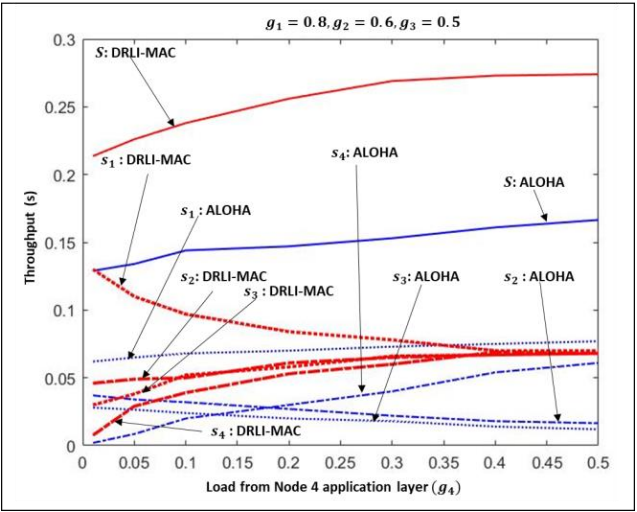

Fig. 9: Performance of DRLI-MAC in a four-node partially network with heterogeneous load

To explore the proposed paradigm's learning ability to handle heterogenous load, experiments were done on a 4-node partially connected topology (see Fig. 8). The load on node-4 is varied while keeping those on nodes 1, 2 and 3 at values higher than the optimal value as indicated by the known benchmark ALOHA protocol. The throughput results are shown in Fig. 9. When compared with ALOHA, it can be seen that the deviation of throughputs among the nodes is significantly reduced, and that is while the maximum throughput is maintained for higher traffic loads. This indicates how learning can take place in the presence of load heterogeneity.

### C. Performance in Large Networks

Fig. 10 shows the load-throughput plots for five-node networks running the proposed MAC learning mechanism. Performance have been evaluated for three different partially connected topologies with maximum two-hop degree of 4, 4 and 3, respectively. It has been observed that learning in these networks attain all three of its distinctive properties, namely, a) attaining optimal throughput as obtained by the known benchmark protocol ALOHA, b) maintaining the optimal throughput at loads higher than what gives rise to the optimal ALOHA throughput, and c) fair distribution of throughput among the nodes in the presence of topological heterogeneity. All these are achieved using MAC learning using only local network information. Also, to be noted that the two latter features are a significant improvement compared to the non-learning based best-known protocol, namely, ALOHA. Similar results are observed in Fig. 11 for larger networks with 8 and 12. The maximum two-hop degrees are 7 for both the networks.

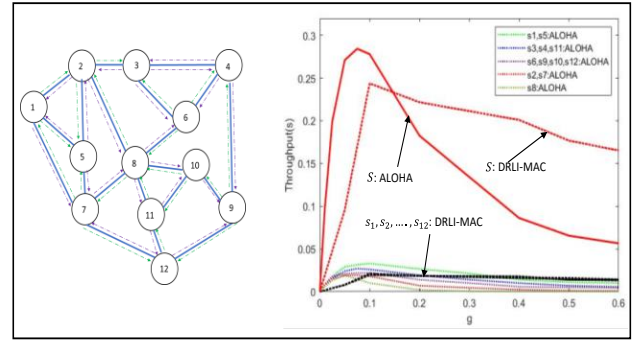

Fig. 12: DRLI-MAC in a 12-node network with 2-hop degree 11

### D. Performance in Denser Networks

Fig. 12 depicts the performance of DRLI-MAC when applied to a dense network with nodal degrees (maximum two-hop degree of 11) that are higher than that of all the networks that were experimented with so far. It can be seen that although the learning-based protocol can achieve throughputs that are higher than those obtained by the benchmark protocol ALOHA, unlike in sparser networks in Figs. 7, 8, 10 and 11, the learning-based strategy cannot sustain that high throughput for larger loading conditions.

Such performance degradation as a function of network density and network size is characterized and presented in Fig. 13. The figure shows that for DRLI-MAC, performance starts degrading when the maximum two hop network degree exceeds around the value 8. However, it is observed from the figure that DRLI-MAC makes the network nodes learn the optimal transmission strategy and give the desired performance even for larger networks as long as the maximum two-hop degree does not exceed 7. This is evident from the plot where the network with 12 nodes gives desired performance for which maximum two-hop degree is 7, but performance degrades for the 12-nodes network with maximum degrees of 9 and 10, respectively. Similar trend is shown in the figure for the 10-nodes network with maximum degrees of 5 and 9, respectively. In short, performance degradation was not observed for larger networks.

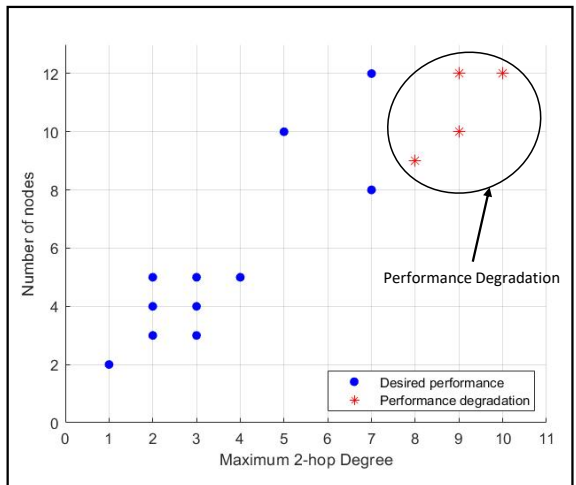

Fig. 13: Performance degradation of DRLI-MAC with increase in network degree and network size

To further investigate the performance degradation in very dense networks, the collision probabilities are recorded against varying 2-hop degrees and load conditions, as shown in Fig. 14. The plots show that the rates of increase of collisions with increasing loads are more prominent in denser networks. This implies that as the network degree increases, the effects of a single node's transmission probabilities on the collisions in its neighborhood become less prominent. Thus, the state space of a learning agent within a node becomes less dependent on the agent's actions. In other words, the system shifts towards being more stateless, thus moving out of the realm of reinforcement learning. To alleviate this, the state space needs to be redefined in a way such that the system does not drift towards being stateless in denser networks. This can be handled using stateless Q-learning, which will be dealt with a future work.

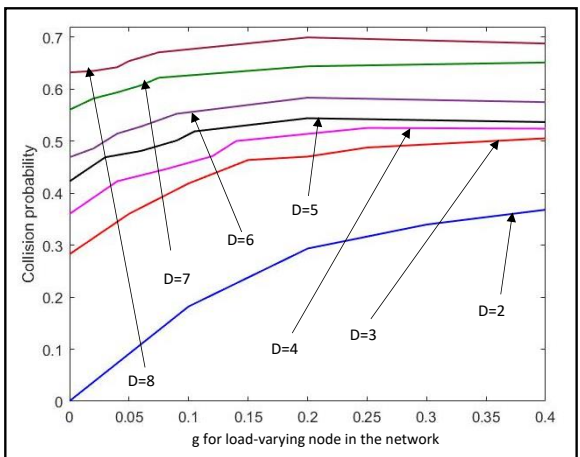

Fig. 14: Variation of Collision probability with maximum 2-hop degree (D) and traffic load (g).

## VI. RELATED WORK

There are several works that deal with designing medium access control protocols using reinforcement learning. The authors in [13] aim at maximizing throughput and minimizing energy requirements in a wireless network using learning-based MAC access. It was implemented by assigning suitable time slots using stateless Q-learning in a slotted time system. Although the solution identifies a valid direction, the presented model is not generalized to work in heterogeneous traffic and network topology scenarios. Moreover, node level access performance is not taken into consideration, thus missing out the fairness performance of the proposed approach. Individual nodes' access performance and heterogeneity are also not considered in [14], in which the energy consumption in wireless sensor networks is tried to be minimized using learning-based access. The approaches in [15, 16], which present Q-learning based adaptive learning for underwater networks, also lack ability to deal with nodal load heterogeneities. Another work in [17] develops a stateless Q learning-based time-slotted MAC protocol that minimizes access collisions for maximizing network throughput. The paper in [18] is designed to work with multiple MAC protocols so that the correct protocols is chosen dynamically depending on various network conditions. Like the prior papers, both [17] and [18] neglects to consider the impacts of load heterogeneity on learning performance.

An RL-based mechanism to increase access fairness in the presence of coexistence between LTE-LAA and WiFi is proposed in [19]. While ensuring fairness, the approach does not target to increase network throughout, and it is also not tested for arbitrary mesh topologies. A scheduler for time-slotted channel hopping, with the goal of minimizing collisions and increasing throughput, is designed using reinforcement learning in [20]. The primary difference of this work with our proposed work is that the protocol in this work does not learn to do load modulation for maintaining good throughput at higher loads. Also, the protocol is not evaluated in sparse networks. The access framework used in [21], although achieves a higher throughput than traditional MAC protocols, it is not able to sustain high throughout at highs loading conditions, as achievable by the mechanism presented in this paper.

A protocol for wireless body area network is designed using reinforcement learning is presented in [7]. This work attempts to increase energy efficiency using a centralized reinforcement learning approach, which is not a practical in many large networks that do not support centralized access arbitration. Maximizing network throughput while improving fairness in a fully connected network using distributed reinforcement learning was proposed in [1]. Although this protocol allows traffic load heterogeneity, the learning approach fails in partially connected network, because it relies on full network-wide information availability, which is not feasible in partially connected topologies. The work presented in this paper addresses this major limitation by incorporating specialized distributed reinforcement learning action, state, and local information sharing abstractions.

## VII. SUMMARY AND CONCLUSIONS

A multi-agent reinforcement learning framework is developed for wireless MAC layer protocol synthesis. The framework allows nodes to learn their transmission strategies so as to reduce the number of collisions. This behavior enables nodes to maintain optimal throughputs for large network loads in a fair manner. The proposed system is shown to achieve desired performance in different heterogeneous scenarios with respect to load distribution and network topologies. The system uses distributed learning, where each node behaves as an independent learner. Distributed learning is useful in large networks where nodes are not expected to keep track of full network-level information and the transmission strategies of all the other nodes. This makes the proposed paradigm generalized for networks with arbitrary mesh topologies. Future work on this topic will include: a) making learning generalized to incorporate stateless RL so that the system does not degrade for dense networks with large node degrees, b) incorporating carrier-sensing in learning for synthesizing robust CSMA family of protocols with better fairness and ability to handle different types of heterogeneity compared to the existing hand-crafted CSMA protocols, and c) incorporating time-slotting to cater to a broader set of network applications that are sensitive to non-throughput performance indices including energy and delay.

## VIII.    REFERENCES

1. Dutta, Hrishikesh, and Subir Biswas. "Towards Multi-agent Reinforcement Learning for Wireless Network Protocol Synthesis." 2021 International Conference on COMmunication Systems & NETworkS (COMSNETS). IEEE, 2021.
2. White III, Chelsea C., and Douglas J. White. "Markov decision processes." European Journal of Operational Research 39.1 (1989): 1-16.
3. Leon-Garcia, Alberto. "Probability, statistics, and random processes for electrical engineering." (2017).
4. Van Otterlo, Martijn, and Marco Wiering. "Reinforcement learning and markov decision processes." Reinforcement Learning. Springer, Berlin, Heidelberg, 2012. 3-42.
5. Watkins, Christopher JCH, and Peter Dayan. "Q-learning." Machine learning 8.3-4 (1992): 279-292.
6. Matignon, Laëtitia, Guillaume J. Laurent, and Nadine Le Fort-Piat. "Hysteretic q-learning: an algorithm for decentralized reinforcement learning in cooperative multi-agent teams." 2007 IEEE/RSJ International Conference on Intelligent Robots and Systems. IEEE, 2007.
7. Xu, Yi-Han, et al. "Reinforcement Learning (RL)-based energy efficient resource allocation for energy harvesting-powered wireless body area network." Sensors 20.1 (2020): 44.
8. Pi, Xurong, Yueming Cai, and Guoliang Yao. "An energy-efficient cooperative MAC protocol with piggybacking for wireless sensor networks." 2011 International Conference on Wireless Communications and Signal Processing (WCSP). IEEE, 2011.
9. Cassandra, Anthony R., Leslie Pack Kaelbling, and Michael L. Littman. "Acting optimally in partially observable stochastic domains." Aaai. Vol. 94. 1994.
10. Kaelbling, Leslie Pack, Michael L. Littman, and Anthony R. Cassandra. "Planning and acting in partially observable stochastic domains." Artificial intelligence 101.1-2 (1998): 99-134.
11. Cassandra, Anthony R., Leslie Pack Kaelbling, and Michael L. Littman. "Acting optimally in partially observable stochastic domains." Aaai. Vol. 94. 1994.
12. Modiano, Eytan. "A novel architecture and medium access control protocol for WDM networks." Optical Networks and Their Applications. Optical Society of America, 1998.
13. Y. Chu, P. D. Mitchell and D. Grace, "ALOHA and Q-Learning based medium access control for Wireless Sensor Networks," 2012 International Symposium on Wireless Communication Systems (ISWCS), Paris, 2012, pp. 511-515, doi: 10.1109/ISWCS.2012.6328420.
14. Savaglio, Claudio, et al. "Lightweight reinforcement learning for energy efficient communications in wireless sensor networks." IEEE Access 7 (2019): 29355-29364.
15. S. H. Park, P. D. Mitchell and D. Grace, "Reinforcement Learning Based MAC Protocol (UW-ALOHA-Q) for Underwater Acoustic Sensor Networks," in IEEE Access, vol. 7, pp. 165531-165542, 2019, doi: 10.1109/ACCESS.2019.2953801.
16. S. H. Park, P. D. Mitchell and D. Grace, "Performance of the ALOHA-Q MAC Protocol for Underwater Acoustic Networks," 2018 International Conference on Computing, Electronics & Communications Engineering (iCCECE), Southend, United Kingdom, 2018, pp. 189-194, doi: 10.1109/iCCECOME.2018.8658631.
17. Lee, Taegyeom, and Ohyun Jo Shin. "CoRL: Collaborative Reinforcement Learning-Based MAC Protocol for IoT Networks." Electronics 9.1 (2020): 143.
18. Yu, Yiding, Taotao Wang, and Soung Chang Liew. "Deep-reinforcement learning multiple access for heterogeneous wireless networks." IEEE Journal on Selected Areas in Communications 37.6 (2019): 1277-1290.
19. Ali, Rashid, et al. "(ReLBT): A Reinforcement learning-enabled listen before talk mechanism for LTE-LAA and Wi-Fi coexistence in IoT." Computer Communications 150 (2020): 498-505.
20. Park, Huiung, et al. "Multi-Agent Reinforcement-Learning-Based Time-Slotted Channel Hopping Medium Access Control Scheduling Scheme." IEEE Access 8 (2020): 139727-139736.
21. Zhenzhen Liu and I. Elhanany, "RL-MAC: A QoS-Aware Reinforcement Learning based MAC Protocol for Wireless Sensor Networks," 2006 IEEE International Conference on Networking, Sensing and Control, Ft. Lauderdale, FL, 2006, pp. 768-773, doi: 10.1109/ICNSC.2006.1673243.